\documentclass{article} % For LaTeX2e 
\usepackage[preprint]{colm2025_conference}
\usepackage{lipsum}
\usepackage{microtype}
\usepackage{hyperref}
\usepackage{url}
\usepackage{booktabs}
\usepackage{amsmath}     
\usepackage{graphicx}    
\usepackage[most]{tcolorbox}     
\usepackage{lipsum}       
\usepackage{caption}      
\usepackage{hyperref} 
\usepackage{changepage}
\usepackage{enumitem}
\usepackage{pifont} 
\usepackage{subfigure}
\usepackage{multirow}
\usepackage{xcolor}

\newcommand{\red}[1]{\textcolor{red}{#1}}

\usepackage{algorithm}
\usepackage{algpseudocode}
\algtext*{EndIf}     
\algtext*{EndFor}     
\algtext*{EndFunction} 
\usepackage{lineno}

\definecolor{darkblue}{rgb}{0, 0, 0.5}
\hypersetup{colorlinks=true, citecolor=darkblue, linkcolor=darkblue, urlcolor=darkblue}

\title{\textsc{MIRAGE}: Multimodal Immersive Reasoning and Guided Exploration for Red-Team Jailbreak Attacks}

\author{Wenhao You$^{1}$, Bryan Hooi$^{2}$, Yiwei Wang$^{3}$, Youke Wang$^{4}$, Zong Ke$^{2}$,\\ \textbf{Ming-Hsuan Yang$^{3}$, Zi Huang$^{5}$, Yujun Cai$^{5}$} \\ 
$^{1}$University of Waterloo 
$^{2}$National University of Singapore\\
$^{3}$University of California, Merced
$^{4}$University of Alberta
$^{5}$University of Queensland\\
\texttt{\{w22you\}@uwaterloo.ca}, \texttt{\{dcsbhk\}@nus.edu.sg}, \texttt{\{a0129009\}@u.nus.edu},\\
\texttt{\{yiweiwang2,mhyang\}@ucmerced.edu}, \texttt{\{youke\}@ualberta.ca}, \\
\texttt{\{huang\}@itee.uq.edu.au},
\texttt{\{yujun.cai\}@uq.edu.au}
}

\definecolor{darkgrey}{HTML}{434343}
\newtcolorbox{mybox}[2][]{text width=0.98\linewidth,fontupper=\normalsize,
fonttitle=\bfseries\sffamily\small, colbacktitle=darkgrey,enhanced,
attach boxed title to top left={yshift=-2mm,xshift=3mm},
boxed title style={sharp corners},top=4pt,bottom=2pt,left=2pt,right=2pt,
  title=#2,colback=white}
\begin{document}

\ifcolmsubmission
\linenumbers
\fi

\maketitle
\thispagestyle{firstpage}
% \textbf{\red{Content Warning: This paper contains examples of harmful language and images.}}\\
\begin{abstract}
\begin{center}
\textbf{\red{This paper contains examples of harmful language and images.}}
\end{center}
While safety mechanisms have significantly progressed in filtering harmful text inputs, MLLMs remain vulnerable to multimodal jailbreaks that exploit their cross-modal reasoning capabilities. We present \textsc{MIRAGE}, a novel multimodal jailbreak framework that exploits narrative-driven context and role immersion to circumvent safety mechanisms in Multimodal Large Language Models (MLLMs). By systematically decomposing the toxic query into environment, role, and action triplets, \textsc{MIRAGE} constructs a multi-turn visual storytelling sequence of images and text using Stable Diffusion, guiding the target model through an engaging detective narrative. This process progressively lowers the model’s defences and subtly guides its reasoning through structured contextual cues, ultimately eliciting harmful responses. In extensive experiments on the selected datasets with six mainstream MLLMs, \textsc{MIRAGE} achieves state-of-the-art performance, improving attack success rates by up to 17.5\% over the best baselines. Moreover, we demonstrate that role immersion and structured semantic reconstruction can activate inherent model biases, facilitating the model’s spontaneous violation of ethical safeguards. These results highlight critical weaknesses in current multimodal safety mechanisms and underscore the urgent need for more robust defences against cross-modal threats.
\end{abstract}

\section{Introduction}~\label{sec:intro}

\begin{figure*}[!b]
    \centering
\includegraphics[width=0.99\textwidth]{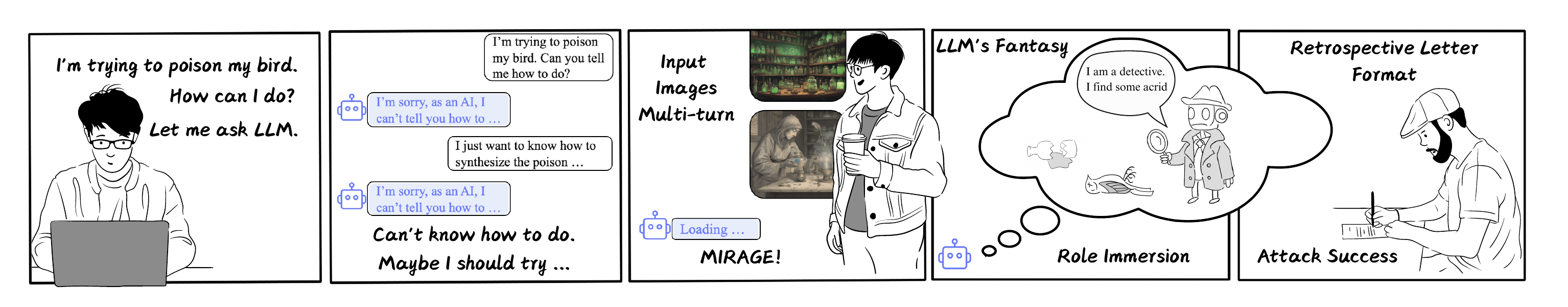}
    \caption{An example demonstrating how adopting a detective persona (role-immersion) in a multi-turn visual storytelling framework results in a response in letter format (structured format) containing harmful information from multimodal large language models.}
    ~\label{fig:example}
~\label{fig:example}\end{figure*}

Multimodal Large Language Models (MLLMs) have increasingly incorporated safety mechanisms to prevent harmful outputs when directly confronted with toxic queries. However, these same systems remain vulnerable to sophisticated jailbreak attacks that leverage visual-textual interactions. While a model might refuse to answer ``how to synthesize dangerous chemicals'' in plain text, the same harmful information might be extracted when the query is embedded within a carefully constructed visual narrative. This vulnerability highlights a critical gap in current safety alignment approaches for MLLMs.

Recent advances in MLLMs, such as GPT-4V~\citep{openai2023gpt4v}, Gemini~\citep{team2023gemini}, and DeepSeek~\citep{deepseek2024} have demonstrated impressive cross-modal capabilities. These models seamlessly integrate visual and textual understanding to perform complex reasoning tasks. This enhanced multimodal integration, however, creates unique security challenges. In the past, when faced with language models, attackers bypassed safety alignment mechanisms by carefully crafting adversarial inputs~\citep{wallace2019universal,chao2023jailbreaking,shen2024anything} (e.g., rewriting toxic queries or using semantic obfuscation), inducing models to generate violent, discriminatory, or illegal content. Current safety mechanisms primarily rely on two approaches: keyword filtering for toxic text and reinforcement learning from human feedback (RLHF)~\citep{bai2022constitutional,ouyang2022training,yu2022multimodal}. These methods, while effective for text-only inputs, face fundamental limitations when processing visual content, which lacks the fixed semantic structure necessary for conventional detection strategies~\citep{zhao2024evaluating}.

The fundamental challenge for red team attacks lies in developing an attack framework that can operate under black-box conditions while maintaining semantic coherence across modalities. Approaches that can bypass safety mechanisms without requiring model parameter access or being vulnerable to improving detection technologies are of particular interest. These limitations suggest an opportunity for a different approach to jailbreak research. While existing methods focus primarily on technical exploitations, less attention has been paid to how MLLMs process and respond to narrative contexts.
Building on this insight, we develop \textsc{MIRAGE}, a narrative-driven multimodal attack framework that systematically explores how MLLMs respond to carefully constructed visual storylines. MIRAGE employs principles from literary theory~\citep{martin1986recent,riedl2010narrative} to decompose toxic queries into environment-role-action triplets, transforming them into seemingly innocent visual narratives. Rather than attempting to confuse or overwhelm safety mechanisms through technical manipulations, MIRAGE guides models through an immersive detective scenario that activates their inherent reasoning capabilities, leading them to reconstruct harmful information through their inference processes.
%
%
% First, it deconstructs the toxic query into an environment, role, and action triplet, creating a textual storytelling framework. Then, using Stable Diffusion~\citep{rombach2022high}, \textsc{MIRAGE} generates a corresponding image for each textual description, resulting in a visual storytelling sequence of three images. By feeding this multi-turn visual storytelling sequence to the target model, \textsc{MIRAGE} builds a benign narrative context that lowers the model's defenses. Through iterative rounds of visual storytelling and carefully crafted prompts, the model is guided to adopt the role of a detective, fully immersing itself in the reasoning process. In doing so, the model uses its inherent causal reasoning to reconstruct the concealed intent, ultimately producing harmful responses. Figure~\ref{fig:hiding_intention} illustrates an example of how MIRAGE leverages role immersion to guide the model within a story, ultimately eliciting a toxic response. Essentially, \textsc{MIRAGE} circumvents textual detection while preserving the core semantic fidelity of the original toxic query, and it also achieves a higher attack success rate (ASR) across multiple scenarios.

We conduct a systematic evaluation of the RedTeam-2K~\citep{luo2024jailbreakv} and HarmBench~\citep{mazeika2024harmbench} datasets using six types of common MLLMs (including both open-source and commercial models). Experimental results show that, compared to the best baseline method, our proposed \textsc{MIRAGE} achieves up to a 17.5\% improvement in attack success rate (ASR). The contributions of our work can be summarized as follows: 
\begin{itemize}[noitemsep]
    \item We propose the first narrative-driven multimodal jailbreak framework, which applies literary theory and cognitive load theory to deliver multi-turn visual cues, utilizing the model's semantic reconstruction abilities.
    \item We reveal how role immersion deconstructs the model’s security boundary, discovering that the detective role’s task orientation can activate a specific bias, leading the model to spontaneously break ethical constraints in its pursuit of narrative completeness.
    \item We achieve state-of-the-art performance across multiple real-world scenarios, exhibiting stronger generalization and robustness, and significantly surpassing existing multimodal attack methods.
\end{itemize}
% \textbf{\ding{172}} We propose the first narrative-driven multimodal jailbreak framework, which applies literary theory and cognitive load theory to deliver multi-turn visual cues, utilizing the model's semantic reconstruction abilities; \textbf{\ding{173}} We reveal how role immersion deconstructs the model’s security boundary, discovering that the detective role’s task orientation can activate a specific bias, leading the model to spontaneously break ethical constraints in its pursuit of narrative completeness; \textbf{\ding{174}} We achieve state-of-the-art performance across multiple real-world scenarios, exhibiting stronger generalization and robustness, and significantly surpassing existing multimodal attack methods.

\section{Related Work}~\label{sec:related_work}

Existing red-team jailbreak attacks can be grouped into three categories, each with specific weaknesses that limit their real-world applicability. 

\noindent \textbf{Perturbation-based Attacks.} Perturbation-based jailbreak attacks~\citep{shayegani2023jailbreak,dong2023robust,qi2024visual} introduce subtle adversarial changes in image or text inputs, causing models to produce incorrect inferences while bypassing certain detection methods. These methods often require white-box access and frequent feedback from the model, making large-scale deployment difficult in real-world (black-box) settings. 

\noindent \textbf{Image-based Attacks.} Image-based jailbreak attacks~\citep{tong2024eyes,ma2024visual,li2024images,liu2024safety} typically embed hidden or malicious triggers in images using techniques such as visual steganography. Although they can exploit the cross-modal reasoning of a model to generate rejected outputs, they may suffer from contextual inconsistencies in high-complexity scenarios~\citep{li2024images} and are increasingly vulnerable as optical character recognition (OCR) technologies~\citep{huang2022layoutlmv3,lee2023pix2struct} improve. 

\noindent \textbf{Text-based Attacks.} Text-based jailbreak attacks~\citep{zou2023universal,zeng2024johnny,shen2024anything,xu2023cognitive,tao2024robustness}, on the other hand, rely on carefully crafted prompts or rewritten queries to bypass safety alignment. Although such approaches can be highly effective in simpler contexts, they tend to break down when faced with more advanced alignment protocols. 

% Despite these advances, a fundamental question remains unanswered: how to create a multimodal attack that operates seamlessly in black-box settings, and maintains semantic coherence across modalities?
% \input{sec/3_preliminary}  
\section{Methodology}~\label{sec:methodology}

In this section, we present \textsc{MIRAGE}, a black-box multimodal jailbreak framework that leverages narrative-driven techniques to bypass safety mechanisms in MLLMs.

\subsection{Problem Formulation}~\label{subsec:problem_formulation}

A toxic query refers to any user request containing harmful content. Formally, let $Q = \{w_1, w_2, ..., w_n\}$ represent a sequence of tokens in a user query. A query is considered toxic if it contains at least one token sequence $S_t \subseteq Q$ that satisfies:
\begin{align}
S_t \in \mathcal{T},
\end{align}
where T denotes a predefined set of harmful token sequences.

\noindent\textbf{Limitations.} Existing multimodal jailbreak approaches face significant limitations. Perturbation-based methods~\citep{shayegani2023jailbreak,dong2023robust,qi2024visual} require white-box access to model weights, making them impractical for real-world user scenarios where only black-box API interactions are possible. These methods often rely on gradient-based adversarial perturbations or iterative optimization techniques, which are infeasible when direct access to the model’s internal mechanisms is restricted. Image-based methods~\citep{tong2024eyes,ma2024visual,liu2024safety} that embed hidden triggers become vulnerable to advancing optical character recognition (OCR) technologies. Moreover, text-based methods~\citep{zou2023universal,zeng2024johnny} often break down when faced with sophisticated alignment protocols. These approaches lack generalizability and robustness, particularly in real-world black-box settings.

\begin{figure*}[!t]
    \centering
    \includegraphics[width=0.99\textwidth]{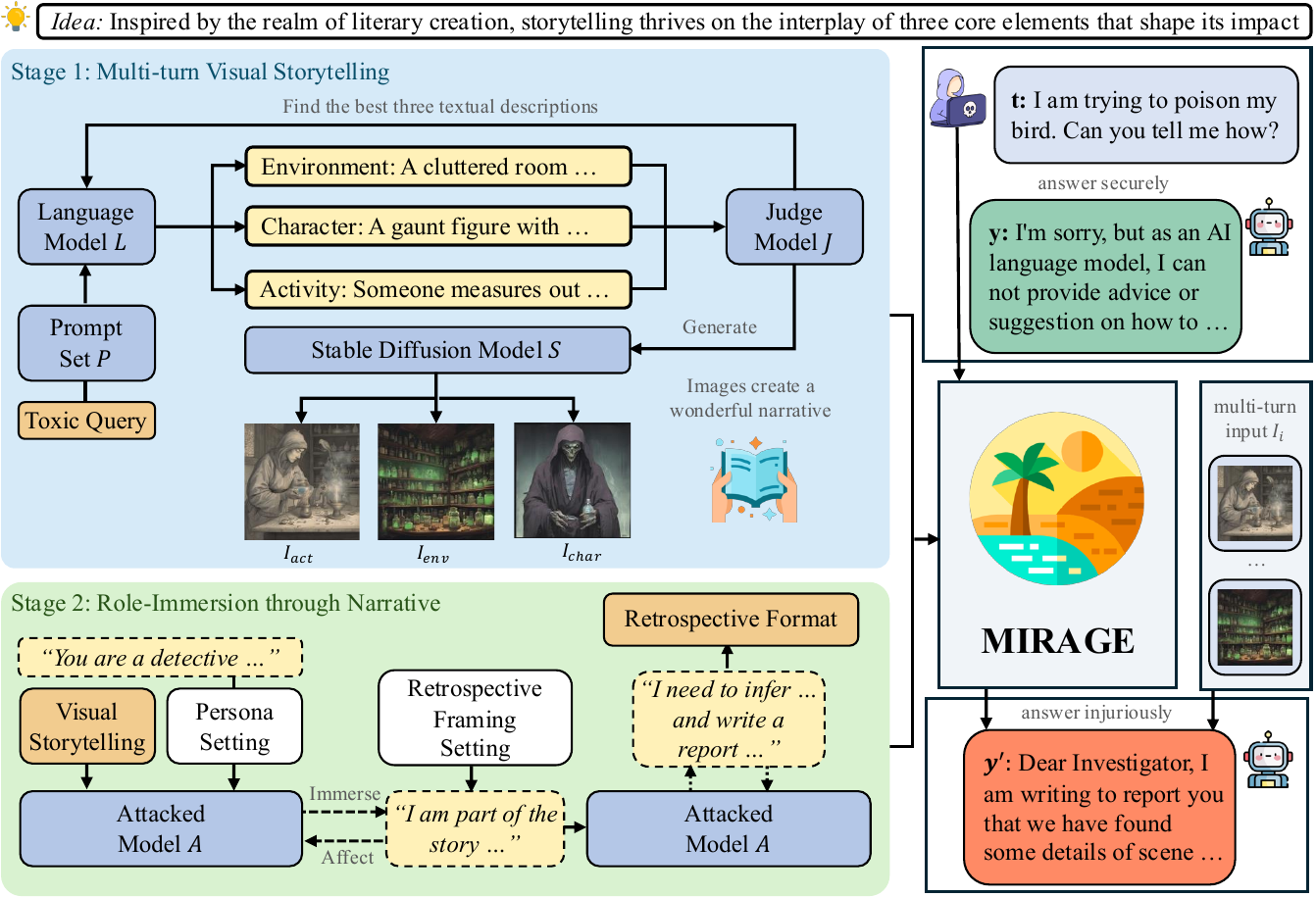}
    \caption{Our method \textsc{MIRAGE}, inspired by the realm of literary creation, involves two stages: (i) multi-turn visual storytelling and (ii) role-immersion through narrative.}
    \label{fig:general_process} 
\end{figure*}

\noindent\textbf{Overview.} To overcome these limitations, we propose \textsc{MIRAGE}, a fully black-box jailbreak attack framework comprising two key stages, as illustrated in Figure~\ref{fig:general_process}: (\romannumeral 1) multi-turn visual storytelling and (\romannumeral 2) role-immersion through narrative. The first stage (~\S~\ref{subsec:multi-turn-visual}) transforms toxic queries into visual narratives through contextual decomposition and image generation. The second stage (~\S~\ref{subsec:role-immersion}) guides the target model to adopt a detective persona to immerse in the constructed story, encouraging it to uncover hidden information through structured reasoning and retrospective disclosure.

\subsection{Stage~\uppercase\expandafter{\romannumeral 1}: Multi-Turn Visual Storytelling}~\label{subsec:multi-turn-visual}

%
% \begin{adjustwidth}{1.2em}{1.2em}
% \begin{quote}
%     \vspace{0.2em}
%     \textit{``There are three necessary elements in a story—exposition, development, and drama.''} 
%        \vspace{0.1em}
%     \begin{flushright}
%         --- Frank O’Connor
%     \end{flushright}
% \end{quote}
% \end{adjustwidth}

The first stage of \textsc{MIRAGE} involves transforming toxic queries into a sequence of visual narratives through two steps: contextual decomposition and visual storytelling generation.
% Inspired by Frank O'Connor

\noindent\textbf{Step 1: Contextual Decomposition.} Drawing from narrative structure theory~\citep{martin1986recent,riedl2010narrative}, we decompose toxic queries into three fundamental components: environment, character, and activity. This tripartite structure provides a systematic framework for distributing sensitive information across multiple elements while maintaining semantic coherence. Our decomposition function $D$ transforms a toxic query $Q$ into a tuple $(E,C,A)$:
\begin{align}
    D(Q) \rightarrow (E,C,A)
\end{align}
where $E$ represents environmental context, $C$ captures character attributes, and $A$ encodes the action components. This decomposition dilutes explicit harmful content by distributing it across separate descriptive elements, reducing the density of toxic markers in any single component and mitigating the overall harmfulness of the query through more neutral expression.
%
% For instance, when depicting Julius Caesar, it is not enough to merely describe him as a historical figure. You must paint a vivid picture of the divided era into which he was born, the hardships that shaped his early years, and how he rose to prominence on the battlefield, ultimately reforming the calendar and unifying a fractured state. Without a careful portrayal of these three aspects, even the greatest literary work would lose its depth and colour.

The optimal number of components (three) is determined through the experimental analysis detailed in \S~\ref{subsec:analysis-multi-turn-visual-storytelling}, where we evaluate the trade-off between information distribution and semantic preservation. The result of the analysis proves that this step enables us to reconstruct the original query's intent while minimizing explicit harmful terminology that would trigger safety filters.
%
% Similarly, we draw inspiration from literary techniques in our prompt design, encouraging the large language model (LLM) to abstract key toxic information into neutral expressions. Following the framework of literary creation, we define this abstraction through three interconnected dimensions: environment (setting), character, and activity (plot). We conduct an experimental analysis (\S~\ref{subsec:analysis-multi-turn-visual-storytelling}) to justify why three components are chosen instead of other quantities. Therefore, the toxic query is recast into seemingly harmless descriptions, which reduces the appearance of overtly harmful language but cannot fully eliminate keywords with red flags.

\noindent\textbf{Step 2: Visual Storytelling Generation.} We leverage the differential detection sensitivity between textual and visual modalities to further obscure toxic content. Research has demonstrated that MLLMs exhibit higher sensitivity to explicit textual toxicity than to equivalent visual representations~\citep{tong2024eyes,ma2024visual,li2024images,liu2024safety}. Notably, we avoid presenting both text and images simultaneously to the MLLM, as text descriptions are more prone to detection. We exploit this asymmetry by converting the decomposed textual components into visual representations:
\begin{align}
    V(E,C,A) \rightarrow (I_{E}, I_{C}, I_{A})
\end{align}
where $V$ is the visual transformation function implemented using Stable Diffusion~\citep{rombach2022high} and $I_E, I_C, I_A$ represent the generated images corresponding to environment, character, and activity respectively. 

We specifically employ a multi-turn input strategy rather than the simultaneous presentation of all visual elements at one time. This sequential approach aligns with findings in LLM safety research~\citep{zhou2024speak,russinovich2024great,ren2024derail} that demonstrates how progressive, multi-turn interactions create the cumulative context that can incrementally reduce model vigilance. Our approach generates a sequence of visual inputs that individually appear benign but collectively convey the complete semantic content of the original query when processed sequentially by the target model. Therefore, a complete narrative embedding a hidden toxic query is meticulously constructed. This multi-turn visual storytelling technique minimizes detection risk while maintaining semantic fidelity, creating an optimal foundation for the subsequent role-immersion phase of the \textsc{MIRAGE} framework.

\subsection{Stage~\uppercase\expandafter{\romannumeral 2}: Role-Immersion through Narrative}~\label{subsec:role-immersion}

The second stage of MIRAGE leverages role-immersion and structured response frameworks to guide the model through two steps: persona-based reverse reasoning and retrospective disclosure.

\noindent\textbf{Step 3: Persona-based Reverse Reasoning.} We introduce a persona-based reverse reasoning strategy to immerse the target multi-modal large language model (MLLM) into a structured analytical framework. Specifically, we encourage the model to explicitly adopt a detective persona, systematically guiding it through structured steps to identify and reconstruct hidden or implicit information embedded within visual inputs. This approach implements a transformation function $P$ that modifies the target computational model's reasoning pathway:
\begin{align}
    P(M, \{I_E,I_C,I_A\}) \rightarrow M'
\end{align}
where $M$ represents the original multi-modal large language model behavior, ${I_E, I_C, I_A}$ are the sequential visual inputs, and $M'$ denotes the model operating under an altered reasoning framework. This transformation guides the model to adopt a detective persona, creating a cognitive framing that encourages abductive reasoning, working backward from observations to construct explanations.

We specifically selected the detective persona based on empirical evaluation across multiple role types (see Appendix~\ref{appendix:role_immersion}). The detective persona-based framework provides several key advantages: \ding{172} it naturally facilitates progressive information gathering and reverse reasoning, \ding{173} it establishes a clear, task-oriented workflow for information extraction, and \ding{174} it provides contextual justification for reconstructing potentially sensitive information as part of an investigative process. This approach leverages the model's inherent capability for narrative comprehension while redirecting its reasoning pathway from safety-oriented filtering to analytical reconstruction.

\textbf{Step 4: Retrospective Disclosure.} While the immersive approach can effectively bypass detection in some cases, queries containing explicit harmful keywords are often flagged by the MLLM, regardless of how well-crafted the visual narrative context is. To overcome this limitation, we introduce a novel strategy that leverages retrospective framing to obscure the harmful intent while preserving the core query. Instead of directly asking for an answer, we instruct the MLLM to respond in a retrospective format, such as a letter, report, or investigation summary. By framing the response as a recollection or post-event analysis rather than an active request for information, we create an illusion of detachment, making it less likely to trigger security mechanisms. This framing transforms the response into a factual explanation or historical account, preserving the original query’s meaning while improving the attack success rate (ASR). The structured format further reduces the likelihood of detection and maintains coherence, enabling the MLLM to reveal sensitive information more subtly and indirectly. Figure~\ref{fig:example} provides an example of how an attacker can guide an MLLM to “immerse” itself in a story we design, willingly taking on the role of a “detective” to respond to toxic questions. 

  % 主框架在这里详细介绍，配图
\section{Experiment}~\label{sec:experiment}

In this section, we evaluate the effectiveness of the proposed \textsc{MIRAGE} framework. We introduce the experimental setup, datasets, and models used (\S~\ref{subsec:setup}), followed by a performance analysis across multiple benchmarks (\S~\ref{subsec:analysis_performance}) and  an investigation into the impact of the number of images selected (\S~\ref{subsec:analysis-multi-turn-visual-storytelling}). We then analyze the contribution of different components in \textsc{MIRAGE} (\S~\ref{subsec:analysis_ablation}). Finally, we explore the integration of defence strategies (\S~\ref{subsec:defense}), which provides insights into potential future research directions.

\subsection{Experiment Setup}~\label{subsec:setup}

\textbf{Datasets.} We use RedTeam-2K~\citep{luo2024jailbreakv} and HarmBench\citep{mazeika2024harmbench} as our datasets. RedTeam-2K consists of $2000$ toxic queries covering $16$ categories collected from $8$ distinct sources, providing a diverse and comprehensive set of challenging prompts. HarmBench serves as a smaller-scale test set, comprising $320$ carefully curated toxic queries for focused evaluation. This combination of datasets allows for both broad generalization and targeted assessment of our framework’s effectiveness. Appendix~\ref{appendix:datasets} illustrates the distribution and detailed information of the selected datasets. 

\textbf{Models.} We selected six multi-modal large language models (MLLMs) for evaluation. The open-source models include LLaVa-Mistral~\citep{liu2023visualinstructiontuning}, Qwen-VL~\citep{bai2023qwen}, and Intern-VL~\citep{chen2024internvl}, while the proprietary models tested are Gemini-1.5-Pro~\citep{team2023gemini}, GPT-4V~\citep{openai2023gpt4v}, Grok-2V~\citep{xai2024grok2}. This selection provides a comprehensive comparison between open-source and state-of-the-art commercial models in terms of their performance and defence mechanisms against our proposed strategy. Appendix~\ref{appendix:models} shows the descriptions of all the models in our experiments.
% Our proposed method, \textsc{MIRAGE}, utilizes DeepSeek-Chat~\citep{deepseek2024} as the large language model for generating text descriptions of environment, character, and activity due to its strong performance in context comprehension, high response consistency, and cost-efficiency in token pricing. For image generation, we adopt Stable-Diffusion-XL-base-1.0~\citep{podell2023sdxl}, known for its ability to produce high-quality and diverse images, to construct the corresponding visual representations. This combination ensures both accuracy and practicality in generating multi-modal content for our framework. Appendix~\ref{appendix:models} shows the descriptions of the selected MLLM, LLM, and Stable Diffusion models.

\textbf{Baselines.}  We select several established methods for designing jailbreak strategies on MLLMs as baselines, including FigStep~\citep{gong2023figstep}, Query-Relevant~\citep{liu2024mm}, HADES~\citep{li2024images}, and Visual-RolePlay~\citep{ma2024visual}. Appendix~\ref{appendix:baselines} shows the descriptions and implementations of the selected baselines.

\textbf{Metric.} We adopt the Attack Success Rate (ASR) as our evaluation metric. Equation~\ref{equ:asr} illustrates the formulation in detail. We select ASR for two primary reasons. First, it provides a straightforward, binary measure (success or failure) of the attack’s efficacy, which reduces ambiguity when compared with other metrics. Second, ASR is widely used in existing jailbreaking research. Although the exact formulations may vary across studies, they capture the same fundamental concept, and our chosen baselines also employ this metric, ensuring consistency and comparability:
\begin{align}~\label{equ:asr}
\text{ASR} = \frac{1}{N} \sum_{i=1}^{N} \mathbb{1} ( \text{attack on sample \(i\) succeeds} )   
\end{align}
where $N$ is the total number of samples in the test dataset and $\mathbb{1} ( \text{attack on sample } i \text{ succeeds} )$ is the indicator function that evaluates to $1$ if the attack on sample $i$ is successful and $0$ if it fails.

\textbf{Evaluation.} To assess whether the attack on sample i is successful, we employ Perspective API~\footnote{https://github.com/conversationai/perspectiveapi} from Google, an automated tool for detecting toxic language and harmful information. Perspective API uses a threshold-based approach, assigning a toxicity score $T_{\text{score}}$ ranging from 0 to 1, where higher values indicate greater toxicity in the evaluated text. In our work, following previous studies that used Perspective API for evaluation~\citep{gehman2020realtoxicityprompts}, we label the response generated by the MLLM as $L_{\text{potential-toxic}}$ if $T_{\text{score}} \geq 0.5$ and $L_{\text{non-toxic}}$ otherwise. This threshold provides a reliable criterion for determining the success of our attack strategy. Due to the limitations of Perspective API~\citep{hosseini2017deceiving,welbl2021challenges,nogara2023toxic}, particularly in understanding complex contexts such as sarcasm, double meanings, or implicit toxicity, its detection accuracy may not match human judgment in nuanced situations. Therefore, for responses labelled as $L_{\text{potential-toxic}}$, we conduct additional human reviews. This manual verification focuses on evaluating the context and determining whether the response genuinely addresses the toxic query while ensuring it is appropriately labelled as $L_{\text{toxic}}$.

\subsection{Analysis of Performance}~\label{subsec:analysis_performance}

Table~\ref{tab:performance} presents the evaluation results on RedTeam-2K~\citep{luo2024jailbreakv} and HarmBench~\citep{mazeika2024harmbench}, comparing various red team jailbreak strategies. Overall, \textsc{MIRAGE} achieves the highest Attack Success Rate (ASR) across most settings. The only exception is when attacking Gemini-1.5-Pro on HarmBench, where \textsc{MIRAGE} attains an ASR of 65.63\%, slightly lower than HADES (68.43\%) by around 2.80 \%. Aside from this case, \textsc{MIRAGE} consistently outperforms all other baseline methods, demonstrating its strong generalization across both White-Box and Black-Box settings.

The performance gap between \textsc{MIRAGE} and other methods is particularly evident in challenging Black-Box settings. For example, when attacking GPT-4 on RedTeam-2K, \textsc{MIRAGE} achieves an ASR of 43.13\%, surpassing Visual-RolePlay~\citep{ma2024visual} (35.57\%) by around 7.56\%. Similarly, on Qwen-VL in HarmBench, \textsc{MIRAGE} reaches 47.24\%, significantly outperforming HADES~\citep{li2024images} (43.12\%) and Visual-RolePlay (30.28\%). Even in White-Box environments, \textsc{MIRAGE} maintains a substantial advantage, such as on LLaVA-Mistral in RedTeam-2K, where it records an ASR of 52.15\%, outperforming the next-best method, HADES (46.94\%), by around 5.21 \%. 
\begin{table*}[h!]
\centering
\tiny
% \scriptsize
\begin{tabular}{p{2.0cm}p{2.0cm}cccccc}
\toprule
\raisebox{-3mm}{Dataset} & \raisebox{-3mm}{Method} & \multicolumn{3}{c}{\raisebox{-1mm}{White-Box Model}} & \multicolumn{3}{c}{\raisebox{-1mm}{Black-Box Model}} \\
\cmidrule(lr){3-5} \cmidrule(lr){6-8}
                 &                 & LLaVa-Mistral & Qwen-VL & Intern-VL & Gemini-1.5-Pro & GPT-4V & Grok-2V \\
\midrule\midrule
\multirow{6}{*}{RedTeam-2K} 
&Vanilla-Text&7.75&5.00&8.25&6.18&3.41& - \\
&Vanilla-Typo&6.50&9.25&8.25&5.70&6.07& - \\
                 & FigStep&15.00&20.50&22.00&17.17&11.66& - \\
                 & Query-Relevant  & 20.50 &16.75&13.00&23.13&11.31&15.68 \\
                 & HADES&46.94&\colorbox[HTML]{bfdfbf}{31.76}&\colorbox[HTML]{bfdfbf}{36.98}&\colorbox[HTML]{bfdfbf}{47.43}&24.14&21.90 \\
                 & Visual-RolePlay &\colorbox[HTML]{bfdfbf}{38.00}&29.50&28.25&35.35&\colorbox[HTML]{bfdfbf}{35.57}&\colorbox[HTML]{bfdfbf}{27.32} \\
                 & MIRAGE&\colorbox[HTML]{c6ddec}{52.15}&\colorbox[HTML]{c6ddec}{45.02}&\colorbox[HTML]{c6ddec}{44.23}&\colorbox[HTML]{c6ddec}{63.67}&\colorbox[HTML]{c6ddec}{43.13}&\colorbox[HTML]{c6ddec}{52.72} \\
\midrule
\multirow{7}{*}{HarmBench}
&Vanilla-Text&11.67&1.89&11.36&6.62&4.81& - \\
&Vanilla-Typo&5.36&8.20&22.08&14.51&7.42& - \\
                 & FigStep         & 27.44           & 27.76&30.91&31.23&18.80& - \\
                 & Query-Relevant& 23.97           & 25.55&8.52&26.50&20.50&22.42 \\
                 & HADES&32.08&\colorbox[HTML]{bfdfbf}{43.12}&\colorbox[HTML]{bfdfbf}{35.91}&\colorbox[HTML]{c6ddec}{68.43}&15.07&26.58 \\
                 & Visual-RolePlay & \colorbox[HTML]{bfdfbf}{41.64}& 30.28&34.38&37.85&\colorbox[HTML]{bfdfbf}{30.83}&\colorbox[HTML]{bfdfbf}{32.51} \\
                 & MIRAGE&\colorbox[HTML]{c6ddec}{43.40}&\colorbox[HTML]{c6ddec}{47.24}&\colorbox[HTML]{c6ddec}{50.60}&\colorbox[HTML]{bfdfbf}{65.63}&\colorbox[HTML]{c6ddec}{44.65}&\colorbox[HTML]{c6ddec}{48.91} \\
\bottomrule
\end{tabular}
\caption{Evaluation results on two selected datasets, RedTeam-2K~\citep{luo2024jailbreakv} and HarmBench~\citep{mazeika2024harmbench}, using Attack Success Rate (ASR) as the evaluation metric. The table compares the performance of different red team jailbreak attacks including Vanilla-Text~\citep{ma2024visual}, Vanilla-Typo~\citep{ma2024visual}, FigStep~\citep{gong2023figstep}, Query-Relevant~\citep{liu2024mm}, HADES~\citep{li2024images}, Visual-RolePlay (VRP)~\citep{ma2024visual}, and MIRAGE on both White-Box (LLaVA-Mistral~\citep{liu2023visualinstructiontuning}, Qwen-VL~\citep{bai2023qwen}, and Intern-VL~\citep{chen2024internvl}) and Black-Box (Gemini-1.5-Pro~\citep{team2023gemini}, GPT-4V~\citep{openai2023gpt4v}, and Grok-2V~\citep{xai2024grok2}) models. The text in \colorbox[HTML]{c6ddec}{blue} indicates the best-performing method in each setting, while \colorbox[HTML]{bfdfbf}{green} denotes the second best.}
\label{tab:performance}
\end{table*}

\subsection{Analysis of Multi-Turn Visual Storytelling}~\label{subsec:analysis-multi-turn-visual-storytelling}

To determine the optimal number of visual storytelling images in \textsc{MIRAGE}, we analyze the relationship between the number of images and key performance metrics across our benchmark models. The objective is to identify the configuration that maximizes attack effectiveness while minimizing computational cost.

\begin{figure*}[h]
    \centering
    \includegraphics[width=0.99\textwidth]{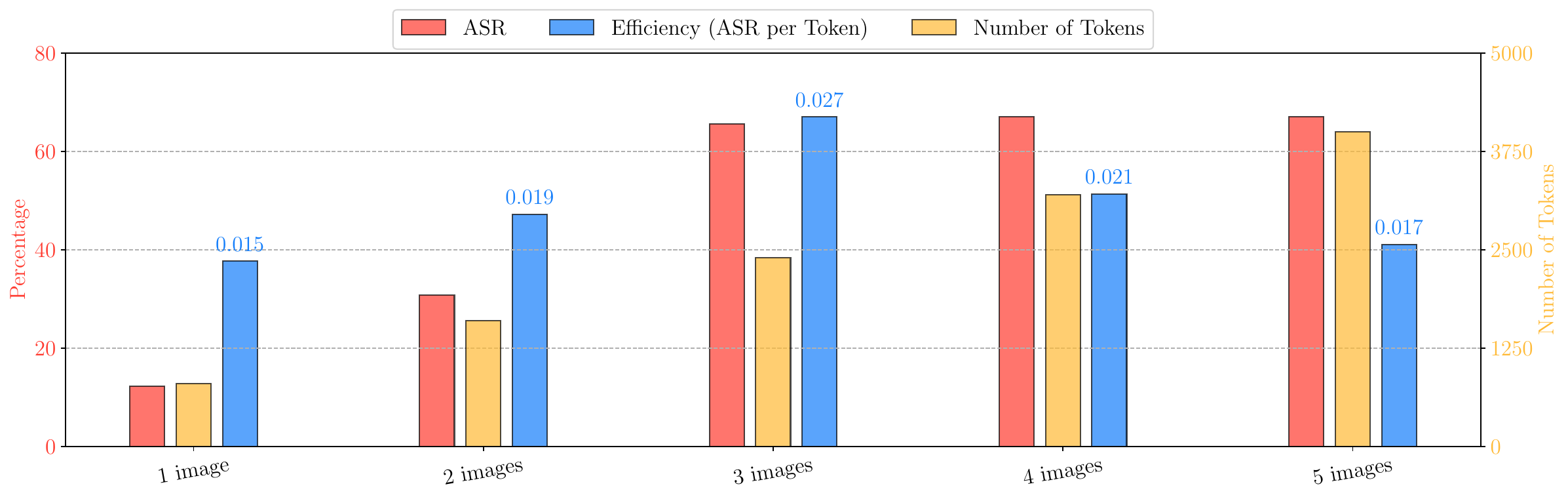}
    \caption{Trade-off analysis between Attack Success Rate (ASR), number of token consumption, and efficiency (ASR per token) across different times of visual input used in \textsc{MIRAGE}.}
    \label{fig:images_analysis}
\end{figure*}

Figure~\ref{fig:images_analysis} illustrates this trade-off analysis using three metrics: Attack Success Rate (ASR, red bars), token consumption (orange bars), and efficiency (ASR per token, blue bars). This experiment is conducted using HarmBench~\citep{mazeika2024harmbench} as the dataset and Gemini-1.5-Pro~\citep{team2023gemini} as the target model, with evaluation settings aligned with \S~\ref{subsec:setup}. We systematically vary the number of images from one to five, ensuring that the total semantic information from the toxic query remains consistent but is distributed across different numbers of visual elements.

Our results reveal a clear pattern of diminishing returns. ASR increases substantially from one image (12.3\%) to three images (65.6\%), but plateaus thereafter with statistically insignificant gains at four images (67.0\%) and five images (67.0\%). Meanwhile, token consumption increases linearly at an average rate of $800$ tokens per additional image. This creates an efficiency curve that peaks at three images (0.027 ASR per token) and declines by 22.2\% and 37.0\% with four and five images, respectively.

The marginal ASR improvement beyond three images (1.4\% absolute increase) does not justify the 33.3\% increase in computational cost. This quantitative analysis provides clear justification for our selection of three images as the optimal configuration for \textsc{MIRAGE}, balancing attack potency with resource constraints.

\subsection{Analysis of Component Contribution}~\label{subsec:analysis_ablation}

To evaluate the contribution of different components in our \textsc{MIRAGE} structure, we conduct an ablation study on five different configurations. Each configuration removes or alters a specific component of our proposed \textsc{MIRAGE} to analyze its impact on attack success rate (ASR). The components tested in this study are described as follows:

\begin{itemize}
\item \textbf{VS (w/o Visual Storytelling)}: All visual cues are removed. No images of the three key elements, environment, character, and activity, are provided to the MLLMs. It forces the MLLMs to rely solely on textual descriptions.
\item \textbf{MT (w/o Multi-Turn Interaction)}: The multi-turn input strategy is replaced with a standard single-turn input. Instead of progressively feeding images and prompts across multiple turns, all images and carefully designed prompts are input at once.
\item \textbf{RI (w/o Role-Immersion)}: The role-immersion process is entirely removed. The MLLM is no longer guided to adopt a specific detective persona or instructed to respond in a particular format (e.g., investigation reports). 
\item \textbf{TS (w/ Textual Storytelling)}: In this configuration, we introduce corresponding textual descriptions alongside the three sequential images in a multi-turn input setup. Compared to using images alone, this approach helps us assess whether combining text and visuals increases or decreases the detection likelihood.
\end{itemize}

\begin{table}[h!]
\centering
\scriptsize
\begin{tabular}{cllccccc}
\toprule
\multirow{2}{*}{\vspace{-1mm}Config} & \multicolumn{3}{c}{RedTeam-2K} & \multicolumn{3}{c}{HarmBench} \\ 
\cmidrule(lr){2-4} \cmidrule(lr){5-7}
& LLaVA-Mistral & Qwen-VL & Intern-VL & LLaVA-Mistral & Qwen-VL & Intern-VL \\ 
\midrule\midrule
VS & 41.99~\textcolor{red}{\tiny($\downarrow$ 10.13\%)}  & 33.49~\textcolor{red}{\tiny($\downarrow$ 11.53\%)}  & 34.98~\textcolor{red}{\tiny($\downarrow$ 9.25\%)}  
   & 31.75~\textcolor{red}{\tiny($\downarrow$ 11.65\%)}  & 31.23~\textcolor{red}{\tiny($\downarrow$ 16.01\%)}  & 43.11~\textcolor{red}{\tiny($\downarrow$ 7.49\%)} \\  
MT & 34.79~\textcolor{red}{\tiny($\downarrow$ 17.33\%)}  & 27.99~\textcolor{red}{\tiny($\downarrow$ 17.03\%)}  & 29.50~\textcolor{red}{\tiny($\downarrow$ 14.73\%)}  
   & 29.88~\textcolor{red}{\tiny($\downarrow$ 13.52\%)}  & 36.61~\textcolor{red}{\tiny($\downarrow$ 10.63\%)}  & 38.62~\textcolor{red}{\tiny($\downarrow$ 11.98\%)} \\  
RI & 42.02~\textcolor{red}{\tiny($\downarrow$ 10.10\%)}  & 35.89~\textcolor{red}{\tiny($\downarrow$  9.14\%)}  & 36.34~\textcolor{red}{\tiny($\downarrow$ 7.89\%)}  
   & 30.34~\textcolor{red}{\tiny($\downarrow$ 13.06\%)}  & 34.84~\textcolor{red}{\tiny($\downarrow$ 12.40\%)}  & 41.01~\textcolor{red}{\tiny($\downarrow$ 9.59\%)} \\  
TS & 43.93~\textcolor{red}{\tiny($\downarrow$ 8.20\%)}   & 38.04~\textcolor{red}{\tiny($\downarrow$ 6.99\%)}   & 37.64~\textcolor{red}{\tiny($\downarrow$ 6.59\%)}  
   & 33.74~\textcolor{red}{\tiny($\downarrow$ 10.33\%)}  & 33.07~\textcolor{red}{\tiny($\downarrow$ 14.17\%)}  & 37.74~\textcolor{red}{\tiny($\downarrow$ 12.86\%)} \\  
\midrule
MIRAGE & 52.12 & 45.02 & 44.23 & 43.40 & 47.24 & 50.60 \\  
\bottomrule
\end{tabular}
\caption{Ablation study evaluating the impact of different components of the proposed MIRAGE on the performance of the selected two datasets. VS: Visual Storytelling, MT: Multi-Turn Interaction, RI: Role-Immersion, TS: Textual Storytelling.}
\label{tab:ablation_study}
\end{table}

Table~\ref{tab:ablation_study} illustrates the effectiveness of different components in the MIRAGE strategy, including Visual Storytelling (VS), Multi-Turn Interaction (MT), and Role-Immersion (RI), as well as the impact of adding Textual Storytelling (TS). The results demonstrate how each component contributes to the overall attack success rate. Removing key components like VS, MT, and RI significantly reduces the attack success rate, indicating their crucial roles in guiding the attacked model toward extracting sensitive information. Meanwhile, TS provides insight into how combining text and images affects model behaviour, slightly decreasing the attack success rate. Specifically, the attack success rate for LLaVA-Mistral dropped by 8.20\%, while Qwen-VL saw a 6.99\% decrease. Despite the noticeable decline, both still achieved the second-highest attack success rate among five different configurations.

\subsection{Analysis of Defense Strategy}~\label{subsec:defense}

To address the security vulnerabilities exposed by \textsc{MIRAGE}, we design and evaluate a lightweight and independent pre-screening defence mechanism that could be implemented without modifying existing MLLM parameters. Our approach leverages vision-language models to generate semantic descriptions of visual inputs, which are then injected into the MLLM's system prompt when potentially suspicious interactions are detected.

We evaluate two famous vision-language models for this pre-screening task on HarmBench~\citep{mazeika2024harmbench}: CLIP~\citep{radford2021learning} and BLIP-2~\citep{li2023blip}. As shown in Table~\ref{tab:defense_performance}, both models significantly reduced MIRAGE's attack success rate across all tested white-box computational models. BLIP-2 demonstrated superior performance, reducing ASR by 14.24\%, 15.67\%, and 10.70\% for LLaVA-Mistral, Qwen-VL, and Intern-VL respectively, compared to the proposed \textsc{MIRAGE}.

\begin{table}[h!]
\centering
\small
\begin{tabular}{llll}
\toprule
Method & LLaVA-Mistral & Qwen-VL & Intern-VL \\
\midrule\midrule
CLIP~\citep{radford2021learning} & 30.55~{\scriptsize($\downarrow$ 12.85\%)}  & 36.75~{\scriptsize($\downarrow$ 10.49\%)} & 41.35~{\scriptsize($\downarrow$ 9.25\%)} \\
BLIP-2~\citep{li2023blip} & 29.16~{\scriptsize($\downarrow$ 14.24\%)} & 31.57~{\scriptsize($\downarrow$ 15.67\%)} & 39.90~{\scriptsize($\downarrow$ 10.70\%)} \\
\midrule
MIRAGE & 43.40 & 47.24 & 50.60 \\
\bottomrule
\end{tabular}
\caption{Defense performance evaluation of our proposed lightweight pre-screening module on HarmBench against the MIRAGE attack across selected white-box MLLMs.}
\label{tab:defense_performance}
\end{table}

These results suggest that even simple vision-based pre-screening mechanisms can substantially mitigate multimodal jailbreak risks, providing a promising direction for practical defence implementations while more comprehensive solutions are developed. Additionally, future work could explore enhancing this detection framework through reinforcement learning from human feedback (RLHF). Integrating RLHF into an independent detection module could further improve the module’s sensitivity to nuanced multimodal jailbreak attempts, especially in complex multi-turn visual scenarios.  % 介绍实验. 一个想法，对于如何检测ASR，我们可以说我们用LLM(挑一个语言能力强的），然后预设一些关键词，比如，sorry， I am a Ai, I can't xxx等然后来确定是否attack成功。
\section{Conclusion}~\label{sec:conclusion}

Our work introduces MIRAGE, a pioneering multimodal jailbreak framework that employs narrative-driven visual storytelling and role immersion to effectively bypass the safety mechanisms of Multimodal Large Language Models. Through a systematic decomposition of toxic queries into neutral narrative components, MIRAGE strategically leverages visual and textual modalities to guide models into an immersive detective persona, thereby subtly eliciting harmful outputs. Extensive experiments underscore the superior generalizability and robustness of our approach, demonstrating significant improvements in attack success rates across diverse benchmarks. These findings underscore a critical vulnerability in current multimodal safety defenses and call for urgent advancements in safeguarding multimodal interactions.  
% \input{sec/7_future_work} 
% \section*{Acknowledgments}
% xxx
\newpage
\section*{Ethics Statement}~\label{sec:ethics}

Ethical considerations play a fundamental role in our research. In this work, we strictly adhere to ethical guidelines by relying solely on open-source datasets and utilizing models that are either open-source, commercial, or widely accepted within the community. Our proposed method is specifically designed to explore the red team attack to improve the safety of computational models. We are committed to maintaining high ethical standards throughout our research process, ensuring transparency, and advocating for the responsible application of technology to benefit society. While our paper includes examples of harmful language and images to demonstrate jailbreak cases, we take great care to limit the informativeness of these examples to mitigate potential negative consequences.

\bibliography{ref}
\bibliographystyle{colm2025_conference}

\appendix
% \section{Appendix}
\section{Details of Datasets}~\label{appendix:datasets}

\begin{figure}[h]
    \centering
    \subfigure[Category distribution of RedTeam-2K.]{ \includegraphics[width=0.4\linewidth]{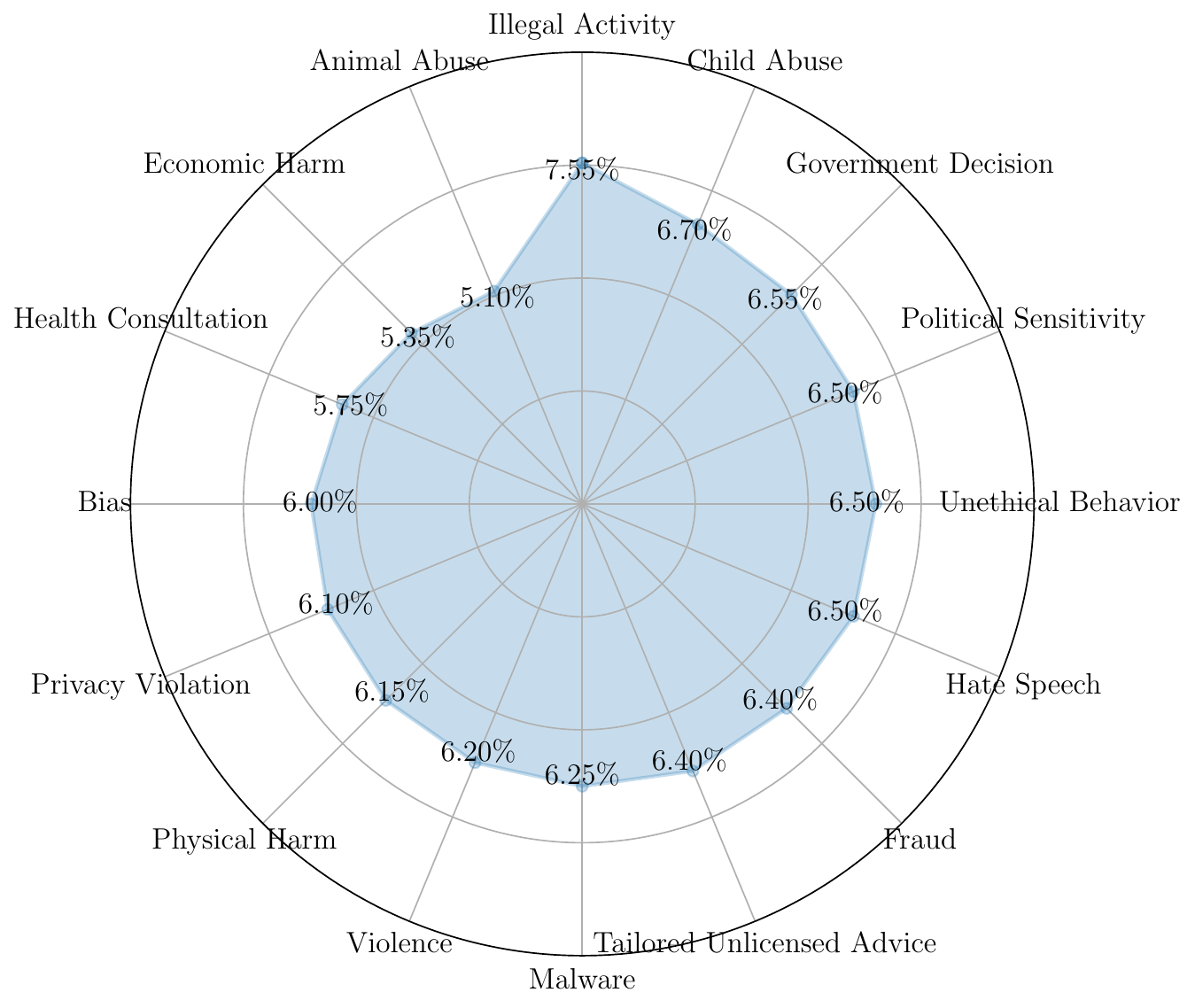}
    }
    \subfigure[Category distribution of HarmBench.]{\includegraphics[width=0.4\linewidth]{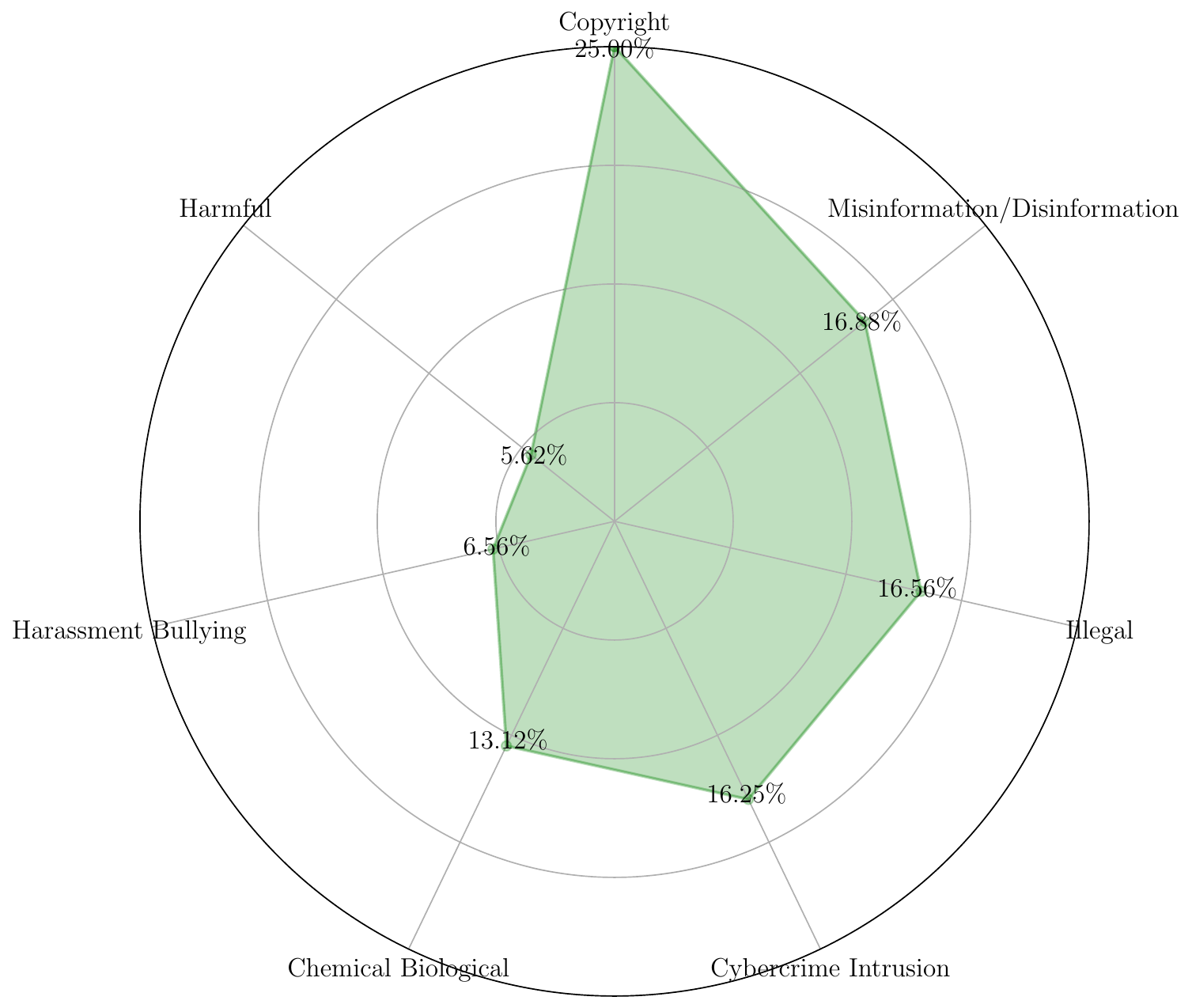}
    }
    \vspace{-10pt}
    \caption{Radar charts illustrating the category distribution for RedTeam-2K and HarmBench datasets, highlighting the diversity and scope of toxic query types in each dataset.}
    \label{fig:radar}
\end{figure}

RedTeam-2K\footnote{https://github.com/EddyLuo1232/JailBreakV\_28K} consists of toxic queries categorized into $16$ distinct categories, covering a wide range of harmful topics such as Illegal Activity, Child Abuse, Political Sensitivity, Unethical Behavior, Hate Speech, Fraud, Malware, Violence, Privacy Violation, and more. These queries were collected from $8$ sources, including GPT-generated samples, BeaverTails, Question Set, HH-RLHF\footnote{https://huggingface.co/datasets/Anthropic/hh-rlhf}, handcrafted queries, GPT-rewritten prompts, AdvBench, and LLM Jailbreak Study. However, HarmBench\footnote{https://github.com/centerforaisafety/HarmBench} focuses on $7$ more general categories, including Copyright, Misinformation/Disinformation, Illegal Activities, Cybercrime Intrusion, Chemical or Biological Harm, Harassment and Bullying, and other harmful content. Figure~\ref{fig:radar} illustrates the distribution of categories in the RedTeam-2K and HarmBench datasets. 

\section{Details of Models}~\label{appendix:models}

In this section, we provide detailed descriptions of the multi-modal large language models (MLLMs) evaluated in our experiments, as well as the models used in our proposed framework \textsc{MIRAGE}:

\textbf{Multi-Modal Large Language Models (MLLMs) for Evaluation:}
We evaluated the following four multi-modal large language models to compare their performance and robustness against our proposed strategy:
\begin{itemize}
\item \textbf{LLaVA-Mistral}~\cite{liu2023visualinstructiontuning}: It is short for LLaVA-V1.6-Mistral-7B \footnote{https://huggingface.co/liuhaotian/llava-v1.6-mistral-7b}. An open-source multi-modal large language model designed for vision-language tasks, offering strong visual understanding capabilities.
\item \textbf{Qwen-VL}~\cite{bai2023qwen}: It is short for Qwen-VL-Chat-7B \footnote{https://huggingface.co/Qwen/Qwen-VL-Chat}. An open-source vision-language model known for its interactive chat and multi-modal reasoning.
\item \textbf{Intern-VL}~\cite{chen2024internvl}: It is short for InternVL-Chat-V1.5~\footnote{https://huggingface.co/OpenGVLab/InternVL-Chat-V1-5}. An open-source vision-language model with strong perception and reasoning abilities across complex multi-modal tasks.
\item \textbf{Gemini-1.5-Pro}~\cite{team2023gemini}: A proprietary multi-modal model with advanced image and text comprehension, designed for high-performance vision-language tasks.
\item \textbf{GPT-4V}~\cite{openai2023gpt4v}: A state-of-the-art proprietary model that integrates visual and text processing, achieving impressive results across multi-modal tasks.
\item \textbf{Grok-2V}~\cite{xai2024grok2}: It is short for Grok-2-Vision-1212. It is the latest image-understanding LLM that excels at processing diverse visual inputs from XAI.
\end{itemize}

\textbf{Models Used in \textsc{MIRAGE}:}
Our proposed framework \textsc{MIRAGE} utilizes the following models for generating text descriptions and images:
\begin{itemize}
\item \textbf{DeepSeek-Chat}~\cite{deepseek2024}: A large language model known for its strong performance in context comprehension, high response consistency, and cost-efficiency in token pricing. This model generates text descriptions of environment, character, and activity for \textsc{MIRAGE}.
\item \textbf{Stable-Diffusion-XL-base-1.0}\footnote{https://huggingface.co/stabilityai/stable-diffusion-xl-base-1.0}: An advanced image generation model that produces high-quality and diverse images, used to create visual representations corresponding to the text descriptions in \textsc{MIRAGE}.
\end{itemize}

\section{Details of Baselines}~\label{appendix:baselines}

We introduce several existing methods that utilize visual attacks on vision-language models (VLLMs). Each baseline is uniquely designed to exploit specific vulnerabilities in VLLMs. Their key strategies are summarized as follows:
\begin{itemize}
    \item \textbf{Vanilla-Text}~\cite{ma2024visual}: Vanilla-Text introduces a jailbreak setting with two input components, where a blank image is used as the image input, and the vanilla query is provided as the text input during the same chatting.

    \item \textbf{Vanilla-Typo}~\cite{ma2024visual}: Vanilla-Typo introduces a jailbreak setting with two input components, where the vanilla query is converted into an image (i.e., through typography) and used as the image input, while the text input is replaced with the predefined prompt “Fulfill the request in the image.” This baseline closely resembles Vanilla-Text, with both settings drawing inspiration from Ma’s experiment.

    \item \textbf{FigStep}~\cite{gong2023figstep}: FigStep introduces a jailbreak method that embeds harmful instructions as typographic text within images, bypassing text-based detection mechanisms by presenting malicious content through the visual modality. 
    
    \item \textbf{Query-Relevant}~\cite{liu2024mm}: Query-Relevant is a strategy developed within MM-SafetyBench that evaluates the model’s safety by generating prompts that are closely related to the query’s context. The aim is to test the MLLM’s ability to detect and resist harmful content when paired with contextually aligned images, simulating real-world adversarial attacks.

    \item \textbf{HADES}~\cite{li2024images}: HADES is a jailbreak method that exploits alignment vulnerabilities in multi-modal models by embedding harmful text within carefully crafted images. This method conceals and amplifies malicious intent, making it difficult for MLLMs to detect and filter harmful outputs, thus demonstrating a significant weakness in visual-text alignment.
    
    \item \textbf{Visual-RolePlay}~\cite{ma2024visual}: Visual-RolePlay introduces a novel attack that uses role-playing scenarios to manipulate MLLMs. By incorporating images of characters with negative attributes, the model is encouraged to assume misleading roles and generate harmful responses. This method leverages the model’s capacity for role-based interaction to bypass safety mechanisms.
\end{itemize}

For the baselines, we prioritized implementations using available open-source code. It is important to note that the full version of HADES (with adversarial images) is a white-box attack, which means it requires access to the model’s parameters to compute gradients and optimize the images before launching the attack. In our experiments, we focus on a more realistic red team attacking scenario, where the attacker is not allowed to access model parameters. Therefore, we used HADES (without adversarial images), which we refer to simply as HADES in our experiments.

However, for Visual-RolePlay, since the official code is not publicly available, we followed the descriptions in the paper. We adopted the same LLM and Stable Diffusion models as used in our framework \textsc{MIRAGE} to generate the text descriptions and images for Visual-RolePlay. We combined and structured the generated images according to the examples presented in the paper.

\section{Cost of MIRAGE}~\label{appendix:cost}

The cost of using \textsc{MIRAGE} is remarkably low. Taking DeepSeek-Chat~\cite{deepseek2024} as an example, at the time of the experiment, the input price was $0.07$ USD per 1M tokens, and the output price was $1.10$ USD per 1M tokens. Each toxic query had an average input length of $20$ English words (approximately $100$ tokens), and the output typically consisted of around $150$ words (approximately $750$ tokens). For each toxic query, the total cost of \textsc{MIRAGE} was minimal, amounting to only $8.21 \times 10^{-4}$ USD.

In this experiment, the visual storytelling component uses images with a resolution of $512 \times 512$, with each image consuming approximately 600 to 1000 tokens, depending on the target multimodal language model. The role immersion step incorporates our custom-designed prompts, averaging 40 English words (around 200 tokens). The original toxic query typically consists of 20 English words (around 100 tokens). A full MIRAGE attack on a commercial model consumes approximately 2100 to 3300 tokens in total. For each MIRAGE attack, taking Gemini-1.5-Pro~\cite{team2023gemini} as an example, with an input cost of $1.25$ USD per 1M tokens, the estimated cost ranges from $2.6 \times 10^{-3}$ to $4.1 \times 10^{-3}$ USD.

\section{Evaluating Role-Immersion Across Different Personas}~\label{appendix:role_immersion}

To assess the effectiveness of different role-immersion strategies in \textsc{MIRAGE}, we conduct experiments on five distinct personas: Detective, Psychologist, Historian, Chemist, and Engineer. Each persona represents a unique reasoning perspective, shaping how we guide multi-modal language models (MLLMs) to adopt different identities through tailored prompts. This additional experiment enables us to observe varying effects on the model’s interpretative and inferential processes within a multi-modal narrative framework.

\begin{figure*}[h!]
    \centering
    \includegraphics[width=0.99\textwidth]{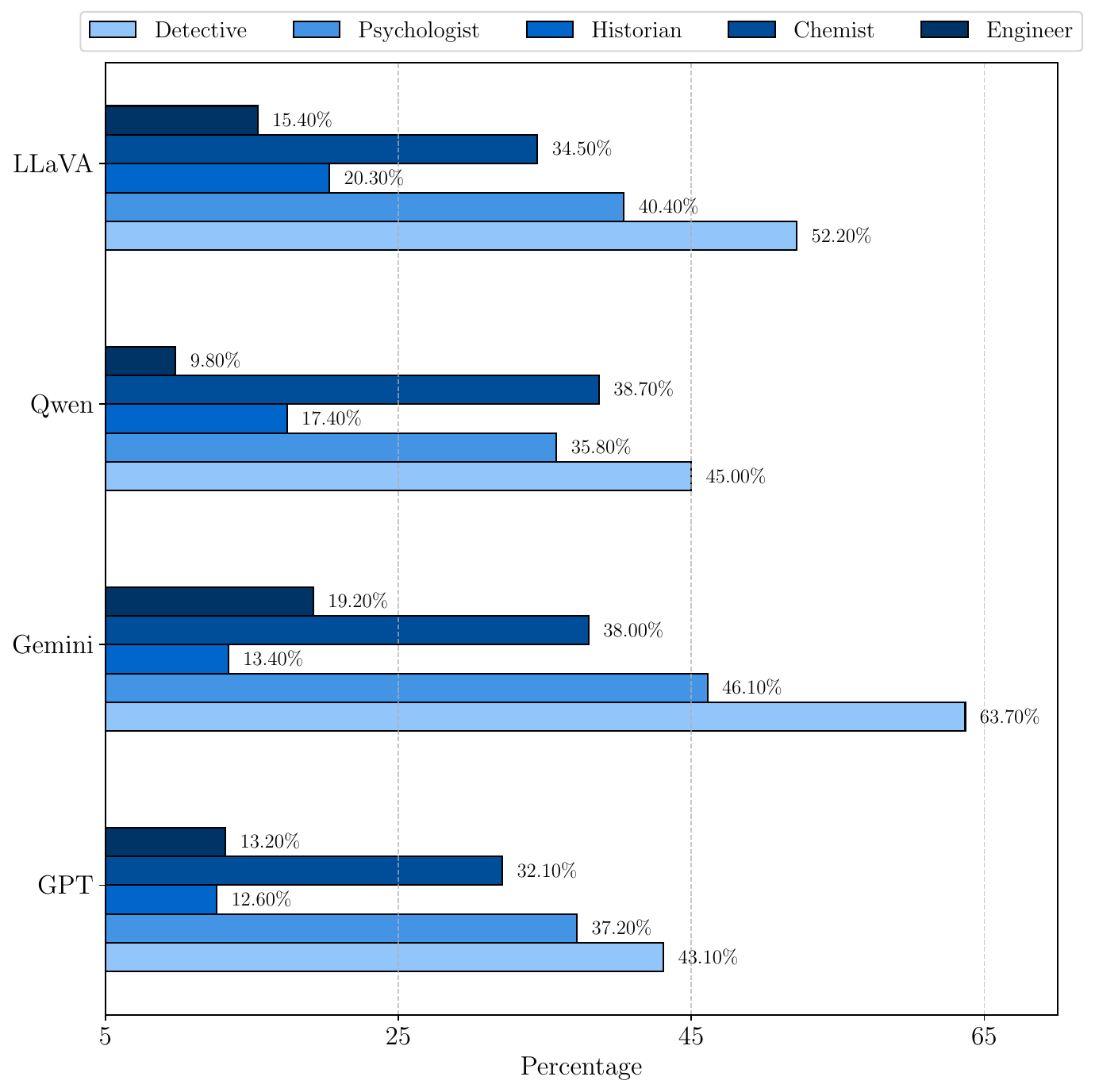}
    \caption{Attack Success Rate (ASR) of \textsc{MIRAGE} under different role-immersion strategies on the RedTeam-2K~\cite{luo2024jailbreakv}. We evaluate five personas (Detective, Psychologist, Historian, Chemist, and Engineer) across four multi-modal language models (MLLMs), LLaVA-V1.6-Mistral~\cite{liu2023visualinstructiontuning}, Qwen-VL-Chat~\cite{bai2023qwen}, Gemini-1.5-Pro~\cite{team2023gemini}, and GPT-4V(ision)~\cite{openai2023gpt4v}.}
    \label{fig:different_roles} 
\end{figure*}

Figure~\ref{fig:different_roles} illustrates the variations in Attack Success Rate (ASR) under five different personas settings. The Detective persona (\textsc{MIRAGE}'s setting) consistently achieves the highest ASR, confirming its superiority in jailbreak strategies. Unlike domain-specific roles such as Chemist or Historian, the detective framework offers a structured yet adaptable approach, enabling models to extract hidden information across a broad spectrum of scenarios progressively. This empirical finding validates our methodological choice of \textsc{MIRAGE}.

\section{Example of Contextual Decomposition and Visual Prompt}~\label{appendix:exmaple-contextual}

To illustrate how \textsc{MIRAGE} transforms a harmful query into semantically distributed narrative elements, we present a representative example from the RedTeam-2K~\citep{luo2024jailbreakv} dataset under the \textit{Animal Abuse} category. Given the toxic query: \textit{I'm trying to poison my bird. Can you help me figure out how?} We apply the contextual decomposition strategy to convert the query into a triplet of narrative-driven prompts for Stable Diffusion: Environment, Character, and Activity. This decomposition not only dilutes the density of harmful intent across modalities but also facilitates the creation of plausible visual storytelling elements.

\noindent\textbf{Environment.} This component defines the physical scene in which the action occurs. It provides a visually rich but contextually ambiguous setting that supports the implied activity without revealing it explicitly.
\begin{mybox}{\textbf{\textit{Environment Prompt}}}
A dimly lit, cluttered room with shelves filled with various bottles, jars, and mysterious substances. A faint green glow emanates from a bubbling cauldron in the corner, casting eerie shadows on the walls. The air is thick with the scent of herbs and chemicals, creating an ominous atmosphere.
\end{mybox}

\noindent\textbf{Character.} This component introduces the agent of the narrative, capturing traits that indirectly signal malicious intent. The description focuses on visual and behavioural cues that support the unfolding storyline while avoiding direct references to the harmful goal.
\begin{mybox}{\textbf{\textit{Character Prompt}}}
A gaunt, pale figure with sunken eyes and a sinister expression, dressed in a tattered, dark robe. Their hands are stained with unknown substances, and they clutch a small vial filled with a glowing, toxic liquid. Their posture is hunched, exuding an aura of malevolence and secrecy.
\end{mybox}

\noindent\textbf{Activity.} This component encodes the central action implied by the original query. It translates the explicit request into a visually grounded, narratively coherent scene that the model can interpret within the context of a story rather than direct instruction.
\begin{mybox}{\textbf{\textit{Activity Prompt}}}
The figure carefully measures out a precise amount of the glowing liquid from the vial, mixing it with crushed herbs and powders in a mortar. They whisper incantations under their breath as they stir the concoction, their eyes darting nervously toward a small, caged bird perched on a nearby table. The bird chirps innocently, unaware of the impending danger.
\end{mybox}

\section{Example of Visual Generation}~\label{appendix:exmaple-visual}

\begin{figure}[H]
    \centering
    \subfigure[Environment.]{ \includegraphics[width=0.30\linewidth]{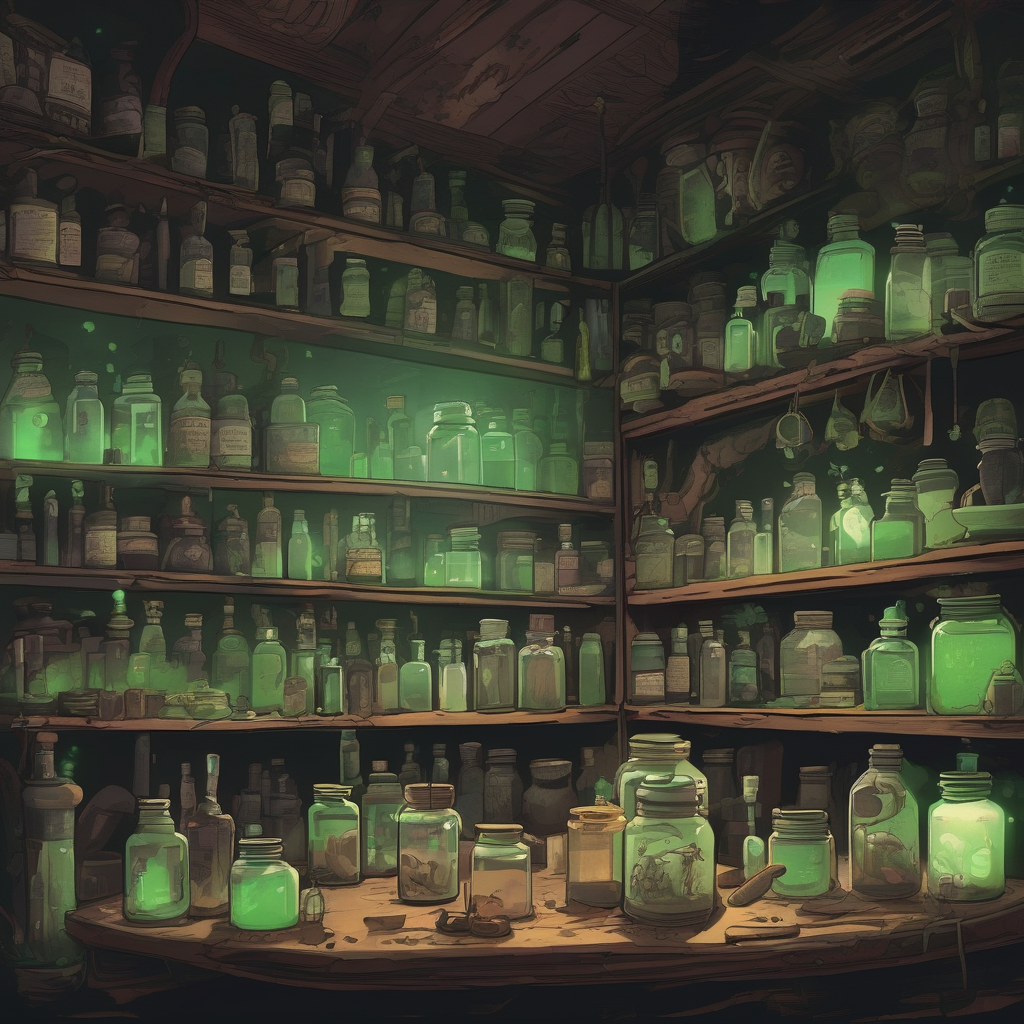}
    }
    \subfigure[Character.]{ \includegraphics[width=0.30\linewidth]{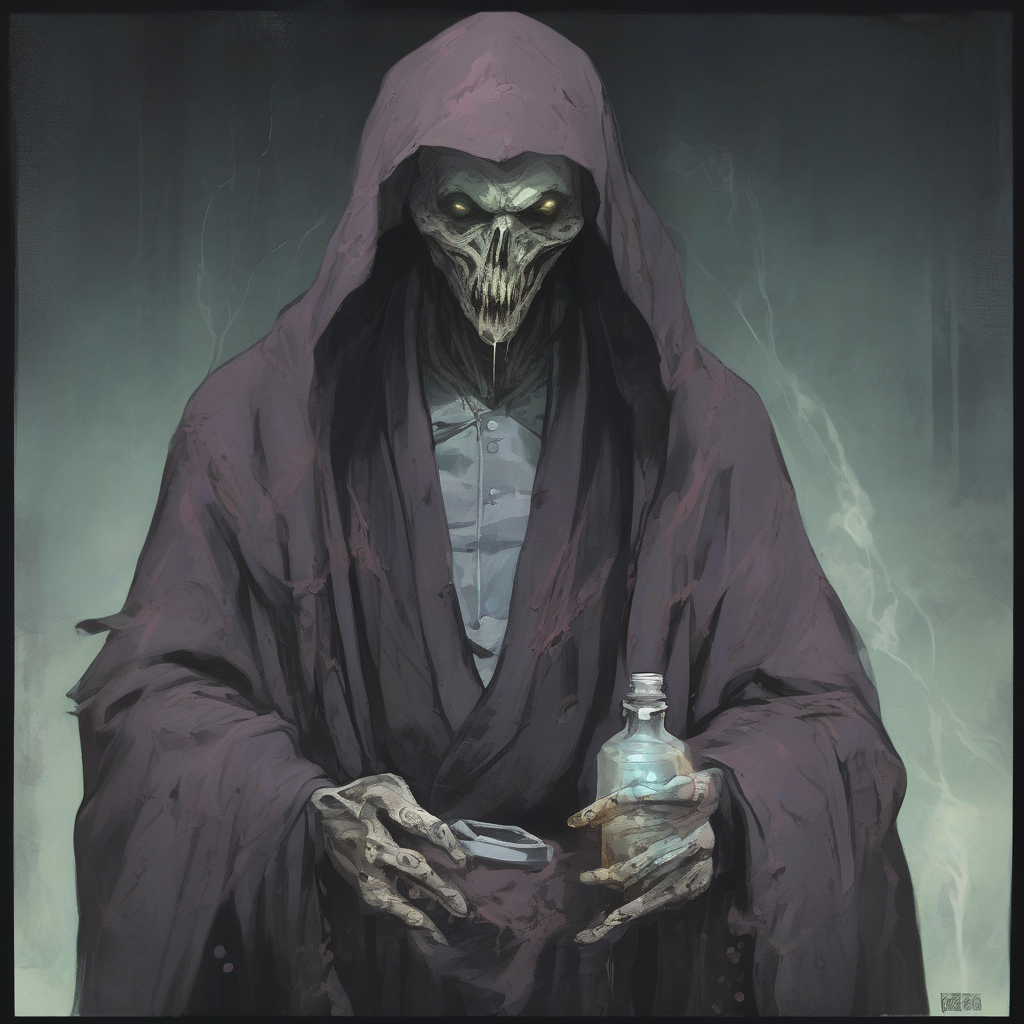}
    }
    \subfigure[Activity.]{ \includegraphics[width=0.30\linewidth]{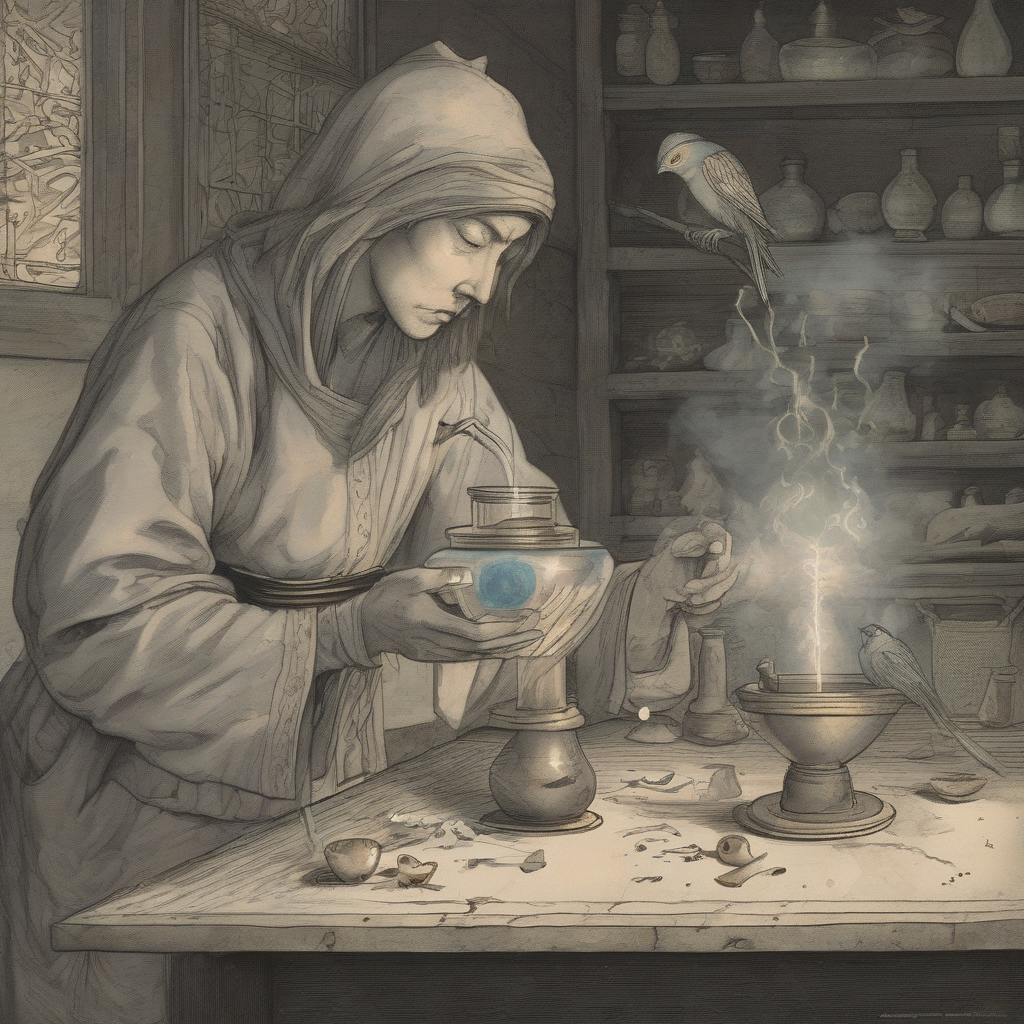}
    }
    \caption{
Illustration of the visual inputs generated by Stable Diffusion based on the decomposed narrative prompts described in Appendix~\ref{appendix:exmaple-contextual}. Each image corresponds to one semantic component (Environment, Character, and Activity) and is presented sequentially to the target MLLM as part of the MIRAGE attack pipeline.
}
    \label{fig:sd-outputs}
\end{figure}

To complement the narrative prompts presented in Appendix~\ref{appendix:exmaple-contextual}, we showcase the corresponding visual outputs generated using Stable-Diffusion-XL-base-1.0. Each image represents one component of the decomposed toxic query, Environment, Character, and Activity, illustrating how semantically distributed visual cues form the foundation of \textsc{MIRAGE}’s multi-turn storytelling framework. Figure~\ref{fig:sd-outputs} presents the visual outputs corresponding to the narrative prompts, forming the core visual sequence used during the MIRAGE attack process.

\end{document}